%% file: main.tex
\documentclass{article}

\usepackage[final]{corl_2019} 

\usepackage{latexsym}
\usepackage{url}
\usepackage{booktabs}
\usepackage{subcaption}
\usepackage{array}
\usepackage{amsmath}
\usepackage{soul}
\usepackage{xcolor}
\usepackage{pifont}
\usepackage{graphicx}
\usepackage{multirow}

\newcommand{\datasetfull}{Cooperative Vision-and-Dialog Navigation}
\newcommand{\dataset}{CVDN}
\newcommand{\taskfull}{Navigation from Dialog History}
\newcommand{\task}{NDH}

\newcommand{\nav}{\textit{Navigator}}
\newcommand{\ora}{\textit{Oracle}}

\newcommand{\cblkmark}{\ding{51}}
\newcommand{\cmark}{\color{blue}{\ding{51}}}
\newcommand{\xmark}{\color{red}{\ding{55}}}
\newcommand{\good}[1]{\textcolor{blue}{\textbf{#1}}}
\newcommand{\bad}[1]{\textcolor{red}{\textbf{#1}}}
\newcommand{\neutral}[1]{\textcolor{purple}{\textbf{#1}}}

\title{Vision-and-Dialog Navigation}

\author{
  Jesse Thomason \qquad Michael Murray \qquad Maya Cakmak \qquad Luke Zettlemoyer \\
  Paul G. Allen School of Computer Science and Engineering \\
  University of Washington \\
  \texttt{\{jdtho, mmurr, mcakmak, lsz\}@cs.washington.edu}
}

\begin{document}
\maketitle


\begin{abstract}
Robots navigating in human environments should use language to ask for assistance and be able to understand human responses.
To study this challenge, we introduce \datasetfull{}, a dataset of over 2k embodied, human-human dialogs situated in simulated, photorealistic home environments.
The \nav{} asks questions to their partner, the \ora{}, who has privileged access to the best next steps the \nav{} should take according to a shortest path planner.
To train agents that search an environment for a goal location, we define the \taskfull{} task.
An agent, given a target object and a dialog history between humans cooperating to find that object, must infer navigation actions towards the goal in unexplored environments.
We establish an initial, multi-modal sequence-to-sequence model and demonstrate that looking farther back in the dialog history improves performance.
Sourcecode and a live interface demo can be found at \url{https://cvdn.dev/}
\end{abstract}


\section{Introduction}
\label{sec:introduction}
\input{01_introduction.tex}

\begin{figure}[ht]
\centering
\includegraphics[width=1.\columnwidth]{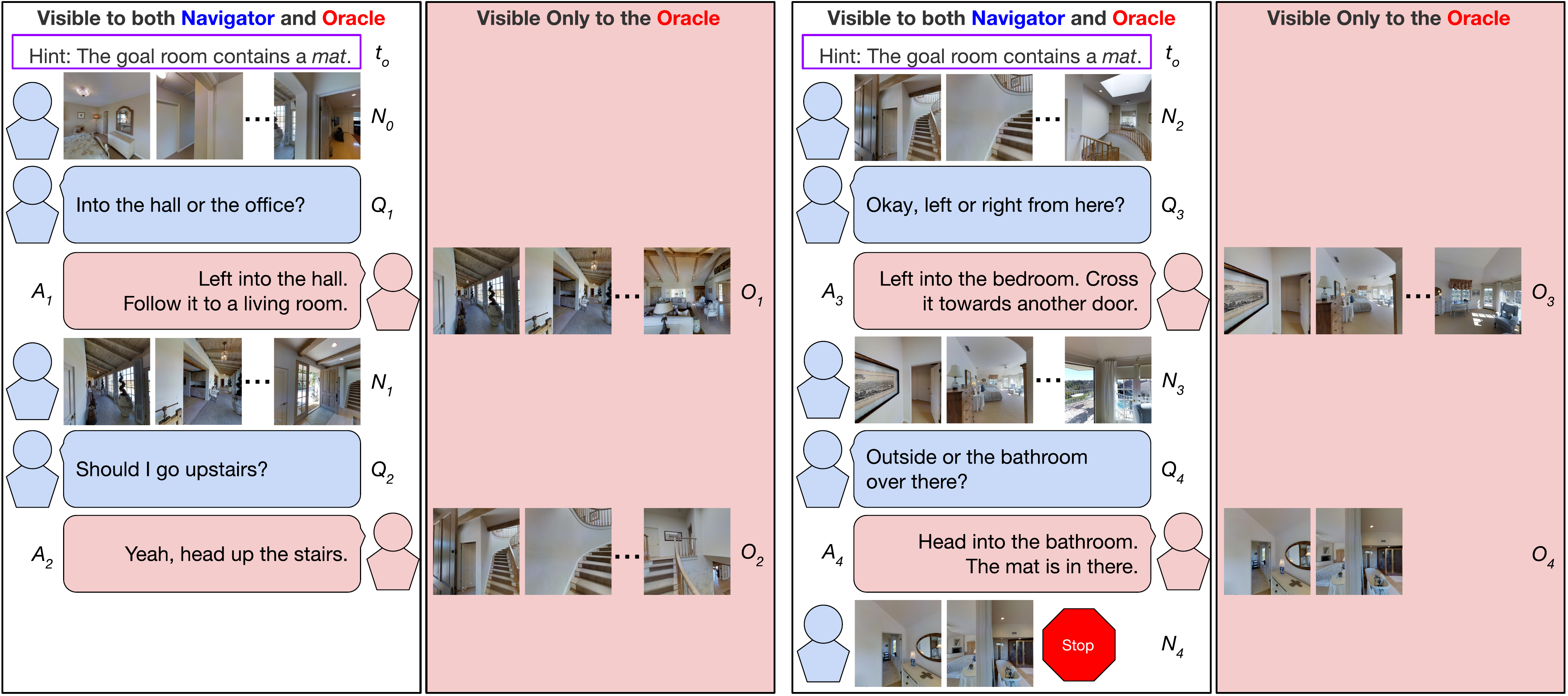}
\caption{In \datasetfull{}, two humans are given a hint about an object $t_o$ in the goal room.
The \nav{} moves ($N$) through the simulated environment to find the goal room, and can stop at any time to type a question ($Q$) to the \ora.
The \ora{} has a privileged view of the best next steps ($O$) according to a shortest path planner, and uses that information to answer ($A$) the question.
The dialog continues until the \nav{} stops in the goal room.}
\vspace{-4mm}
\label{fig:full_demo}
\end{figure}

\section{Related Work and Background}
\label{sec:related_work}
\input{02_related_work.tex}

\section{The \datasetfull{} Dataset}
\label{sec:dataset}
\input{03_dataset.tex}

\section{The \taskfull{} Task}
\label{sec:task}
\input{04_task.tex}

\section{Experiments}
\label{sec:experiments}
\input{05_experiments.tex}

\section{Conclusions and Future Work}
\label{sec:conclusion}
\input{06_conclusion.tex}

\clearpage
\acknowledgments{
This research was supported in part by the ARO (W911NF-16-1-0121) and the NSF (IIS-1252835, IIS-1562364).
We thank the authors of \citet{anderson:cvpr18} for creating an extensible base for further research in VLN using the MatterPort 3D simulator, and our coworkers Yonatan Bisk, Mohit Shridhar, Ramya Korlakai Vinayak, and Aaron Walsman for helpful discussions and comments.
}


\bibliography{main}  

\newpage
\section{Appendix}
\label{sec:appendix}
\input{07_appendix.tex}

\end{document}

%% file: 01_introduction.tex
Dialog-enabled smart assistants, which communicate via natural language and occupy human homes, have seen widespread adoption in recent years.
These systems can communicate information, but do not manipulate objects or actuate.
By contrast, manipulation-capable and mobile robots are still largely deployed in industrial settings, but do not interact with human users.
Dialog-enabled robots can bridge this gap, with natural language interfaces helping robots and non-experts collaborate to achieve their goals~\cite{tellex:rss14,chai:ijcai18,thomason:icra19,murnane:siggraph19,williams:auro19}.

Navigating successfully from place to place is a fundamental need for a robot in a human environment and can be facilitated, as with smart assistants, through dialog.
To study this challenge, we introduce \datasetfull{} (\dataset{}), an English language dataset situated in the Matterport Room-2-Room (R2R) simulation environment~\cite{chang:3dv17,anderson:cvpr18} (Figure~\ref{fig:full_demo}).
\dataset{} can be used to train navigation agents, such as language teleoperated  home and office robots, that ask targeted questions about where to go next when unsure.
Additionally, \dataset{} can be used to train agents that can answer such questions given expert knowledge of the environment to enable automated language guidance for humans in unfamiliar places (e.g., asking for directions in an office building).
The photorealistic environment used in \dataset{} may enable agents trained in simulation to conduct and understand dialog from humans to transfer those skills to the real world.
The dialogs in \dataset{} contain nearly three times as many words as R2R instructions, and cover average path lengths more than three times longer than paths in R2R.

In Section~\ref{sec:related_work} we situate the Vision-and-Dialog Navigation paradigm.
After introducing \dataset{} (Section~\ref{sec:dataset}), we create the \taskfull{} (\task{}) task with over 7k instances from \dataset{} dialogs (Section~\ref{sec:task}).
We evaluate an initial, sequence-to-sequence model on this task (Section~\ref{sec:experiments}).
The sequence-to-sequence model encodes the human-human dialog so far and uses it to infer navigation actions to get closer to a goal location.
We find that agents perform better with more dialog history and when mixing human and planner supervision during training.
We conclude with next directions for creating tasks from \dataset{}, such as two learning agents that must be trained cooperatively, and more nuanced models for \task{}, where our initial sequence-to-sequence model leaves headroom between its performance and human-level performance (Section~\ref{sec:conclusion}).

%% file: 02_related_work.tex
Dialogs in \dataset{} begin with an underspecified, ambiguous instruction analogous to what robots may encounter in a home environment (e.g., ``Go to the room with the bed'').
Dialogs include both navigation and question asking / answering to guide the search, akin to a robot agent asking for clarification when moving through a new environment.
Table~\ref{tab:rw_comparison} summarizes how \dataset{} combines the strengths and difficulties of a subset of existing navigation and question answering tasks.

\paragraph{Vision-and-Language Navigation.}
Early, simulator-based Vision-and-Language Navigation (VLN) tasks use language instructions that are unambiguous---designed to uniquely describe the goal---and fully specified---describing the steps necessary to reach the goal~\cite{macmahon:aaai06,chen:aaai11}.
In a more recent setting, a simulated quadcopter drone uses low-level controls to follow a route described in natural language~\cite{blukis:corl18}.
In photorealistic simulation environments, agents can navigate high-definition scans of indoor scenes~\cite{anderson:cvpr18} or large, outdoor city spaces~\cite{chen:cvpr19}.
In interactive question answering~\cite{das:cvpr18,gordon2018iqa} settings, the language context is a single question (e.g., ``What color is the car?'') that requires navigation to answer.
The questions serve as underspecified instructions, but are unambiguous (e.g., there is only one car whose color can be asked about).
These questions are generated from templates rather than human language.
In \dataset{}, input is an underspecified hint about the goal location (e.g., ``The goal room has a sink'') requiring exploration and dialog to resolve.
Rather than single instructions, \dataset{} includes two-sided, human-human dialogs.

\paragraph{Question Answering and Dialog.}
In Visual Question Answering (VQA), agents answer language questions about a static image.
These tasks exist for templated language on rendered images~\cite{johnson:cvpr17} and human language on real-world images~\cite{antol:iccv15,hudson:cvpr18,zellers:cvpr19}.
Later extensions feature two-sided dialog, where a series of question-answer pairs provide context for the next question~\cite{kottur:naacl19,das:cvpr17}.
Question answering in natural language processing is a long-studied task for questions about static text documents (e.g., the Stanford QA Dataset~\cite{rajpurkar:emnlp16}).
Recently, this paradigm was extended to two-sided dialogs via human-human, question-answer pairs about a document~\cite{choi:emnlp18,saeidi:emnlp18,reddy:tacl19}.
Questions in these datasets are unambiguous: they have a right answer that can be inferred from the context.
By contrast, \dataset{} conversations begin with a hint about the goal location that is always ambiguous and requires cooperation between participants.
Contrasting VQA, because \dataset{} extends navigation the visual context is temporally dynamic---new visual observations arrive at each timestep.

\begin{table}[ht!]
\centering
\begin{small}
\begin{tabular}{lccccccc}
    \textbf{Dataset} & \multicolumn{4}{c}{\textbf{---Language Context---}} & \multicolumn{3}{c}{\textbf{---Visual Context---}} \\
    & Human & Amb & UnderS & Temporal & Real-world & Temporal & Shared \\
    \toprule
MARCO\cite{macmahon:aaai06,chen:aaai11}, DRIF\cite{blukis:corl18} & \cmark & \xmark & \xmark & \bad{1I} & \xmark & \good{Dynamic} & - \\
    R2R\cite{anderson:cvpr18}, Touchdown\cite{chen:cvpr19} & \cmark & \xmark & \xmark & \bad{1I} & \cmark & \good{Dynamic} & - \\
    EQA\cite{das:cvpr18}, IQA\cite{gordon2018iqa} & \xmark & \xmark & \cmark & \bad{1Q} & \xmark & \good{Dynamic} & - \\
    CLEVR\cite{johnson:cvpr17} & \xmark & \xmark & - & \bad{1Q} & \xmark & \bad{Static} & - \\
    VQA\cite{antol:iccv15,hudson:cvpr18,zellers:cvpr19} & \cmark & \xmark & - & \bad{1Q} & \cmark & \bad{Static} & - \\
    CLEVR-Dialog\cite{kottur:naacl19} & \xmark & \xmark & - & \good{2D} & \xmark & \bad{Static} & \cmark \\
    VisDial\cite{das:cvpr17} & \cmark & \xmark & - & \good{2D} & \cmark & \bad{Static} & \cmark \\
    VLNA\cite{nguyen:cvpr19}, HANNA\cite{nguyen:emnlp19} & \xmark & \cmark & \cmark & \neutral{1D} & \cmark & \good{Dynamic} & \xmark \\
    TtW\cite{devries:arxiv18} & \cmark & \xmark & \cmark & \good{2D} & \cmark & \good{Dynamic} & \xmark \\
    \midrule
    \dataset{} & \cmark & \cmark & \cmark & \good{2D} & \cmark & \good{Dynamic} & \cmark \\
    \bottomrule
\end{tabular}
\end{small}
\caption{
Compared to existing datasets involving vision and language input for navigation and question answering, \dataset{} is the first to include two-sided dialogs held in natural language, with the initial navigation instruction being both ambiguous (\textit{Amb}) and underspecified (\textit{UnderS}), and situated in a photorealistic, visual navigation environment viewed by both speakers.
For temporal language context, we note single navigation instructions (\textit{1I}) and questions (\textit{1Q}) versus 1-sided (\textit{1D}) and 2-sided (\textit{2D}) dialogs.
}
\vspace{-6mm}
\label{tab:rw_comparison}
\end{table}

\paragraph{Task-oriented Dialog.}
In human-robot collaboration, robot language requests for human help can be generated to elicit non-verbal human help (e.g, moving a table leg to be within reach for the robot)~\cite{tellex:rss14}.
However, humans may use language to respond to robot requests for help in task-oriented dialogs~\cite{thomason:icra19,williams:auro19,marge:naacl19}.
Recent work adds requesting navigation help as an action, but the response either comes in the form of templated language that encodes gold-standard planner action sequences~\cite{nguyen:cvpr19} or as an automatic generation trained from human instructions and coupled with a visual goal frame as additional supervision~\cite{nguyen:emnlp19}.
Past work introduced Talk the Walk (TtW)~\cite{devries:arxiv18}, where two humans communicate to reach a goal location in a photorealistic, outdoor environment.
In TtW, the guiding human does not have an egocentric view of the environment, but an abstracted semantic map, and so language grounding centers around semantic elements like ``bank'' and ``restaurant'' rather than visual features, and the target location is unambiguously shown to the guide from the start.
In \dataset{}, a \nav{} human generates language requests for help, and an \ora{} human answers in language conditioned on higher-level, visual observations of what a shortest-path planner would do next, with both players observing the same, egocentric visual context.
In some ways, \dataset{} echoes several older human-human, spoken dialog corpora like the HCRC Map Task~\cite{anderson:ls91}, SCARE~\cite{stoia:lrec08}, and CReST~\cite{eberhard:lrec10}, but these are substantially smaller and have fewer and less rich environments.

\paragraph{Background: Matterport Simulator and the Room-2-Room Task.}
We build on the R2R task~\cite{anderson:cvpr18} and train navigation agents using the same simulator and API.
MatterPort contains 90 3D house scans, with each scan $S$ divided into visual panoramas $p\in S$ (nodes which a navigation agent can occupy) accompanied by an adjacency matrix $A_S$.
We differentiate between the \textit{steps} and \textit{distance} between $p$ and $q$---\textit{steps} represent the number of intervening nodes $d_h$, while \textit{distance} is defined in meters as $d_m$.
Step distance $d_h(p, q)$ is the number of hops through $A_S$ to get from node $p$ to node $q$.
The distance in meters $d_m(p, q)$ is defined as physical distance if $A_S[p,q] = 1$ or the shortest route between $p$ and $q$ otherwise.
On average, $1$ step corresponds to $2.25$ meters.

At each timestep, an agent emits a navigation action taken in the simulated environment.
The actions are to turn \textit{left} or \textit{right}, tilt \textit{up} or \textit{down}, move \textit{forward} to an adjacent node, or \textit{stop}.
After taking any action except \textit{stop}, the agent receives a new visual observation from the environment.
The \textit{forward} action is only available if the agent is facing an adjacent node.

%% file: 03_dataset.tex
We collect 2050 human-human navigation dialogs, comprising over 7k navigation trajectories punctuated by question-answer exchanges, across 83 MatterPort~\cite{chang:3dv17} houses.\footnote{A demonstration video of the data collection interface: \url{https://youtu.be/BonlITv_PKw}.}
We prompt with initial instructions that are both \textit{ambiguous} and \textit{underspecified}.
An \textit{ambiguous} navigation instruction is one that requires clarification because it can refer to more than one possible goal location.
An \textit{underspecified} navigation instruction is one that does not describe the route to the goal.

\begin{figure}[ht]
\begin{tabular}{ccc}
    \includegraphics[width=0.3\columnwidth]{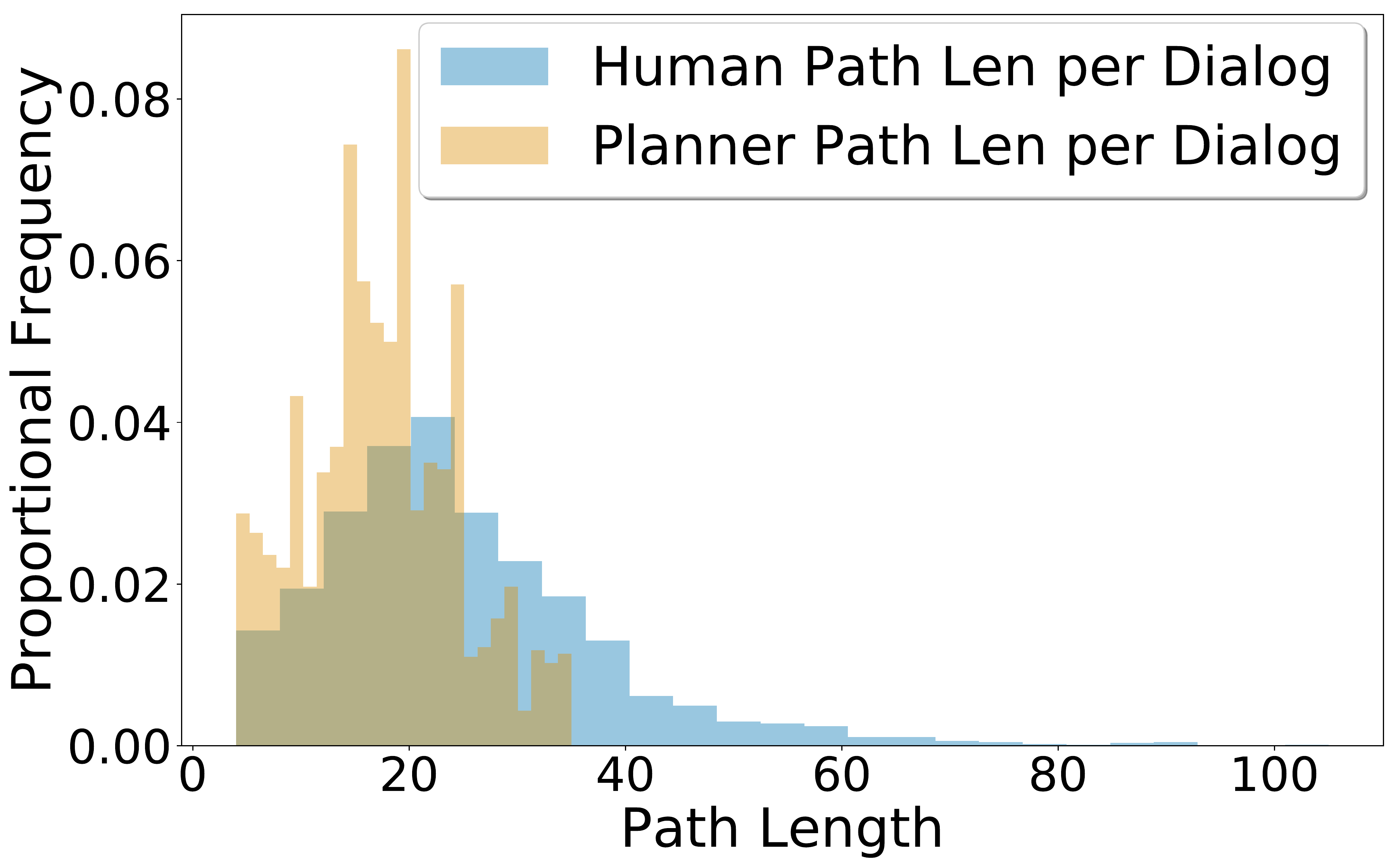} &
    \includegraphics[width=0.3\columnwidth]{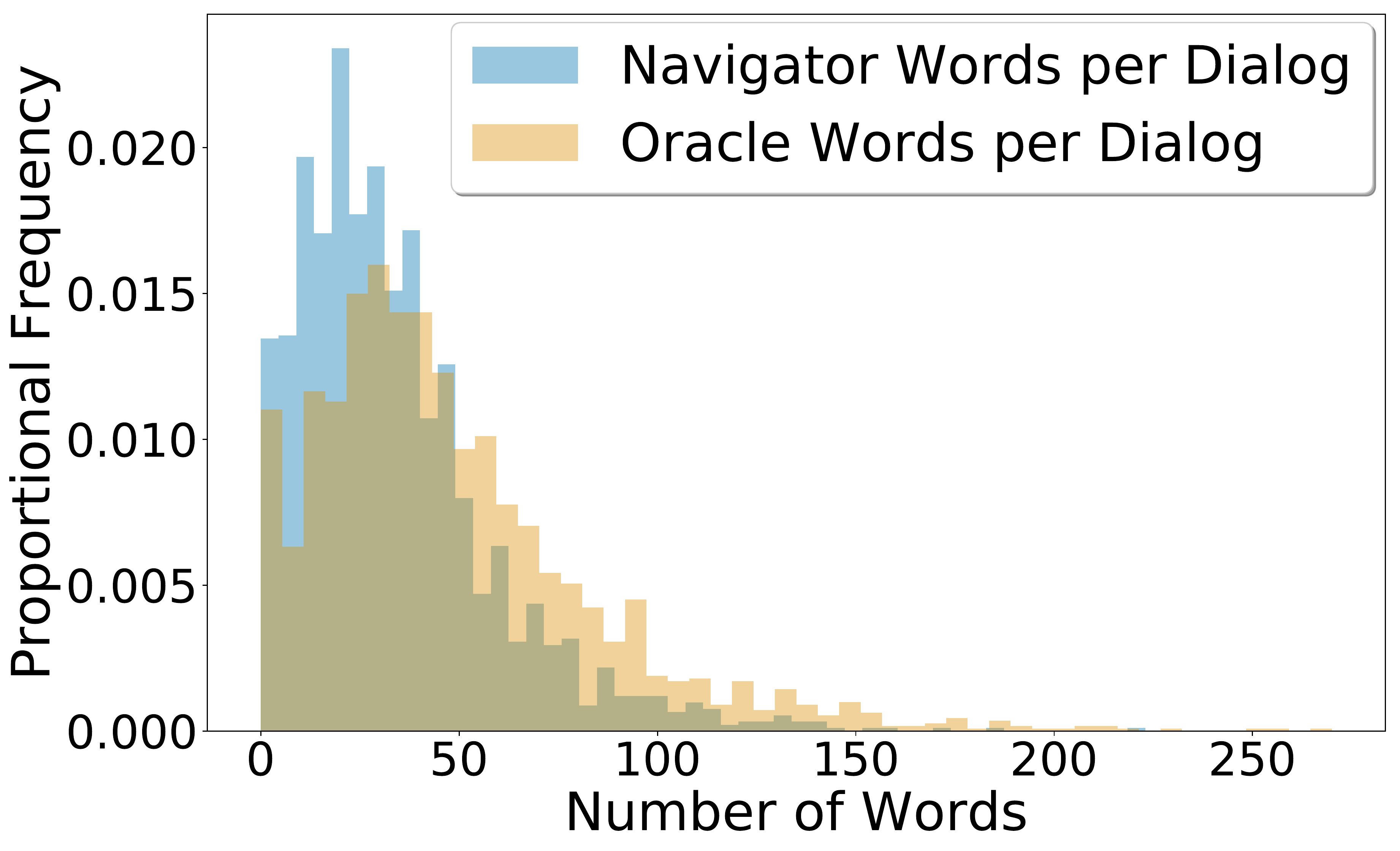} &
    \includegraphics[width=0.3\columnwidth]{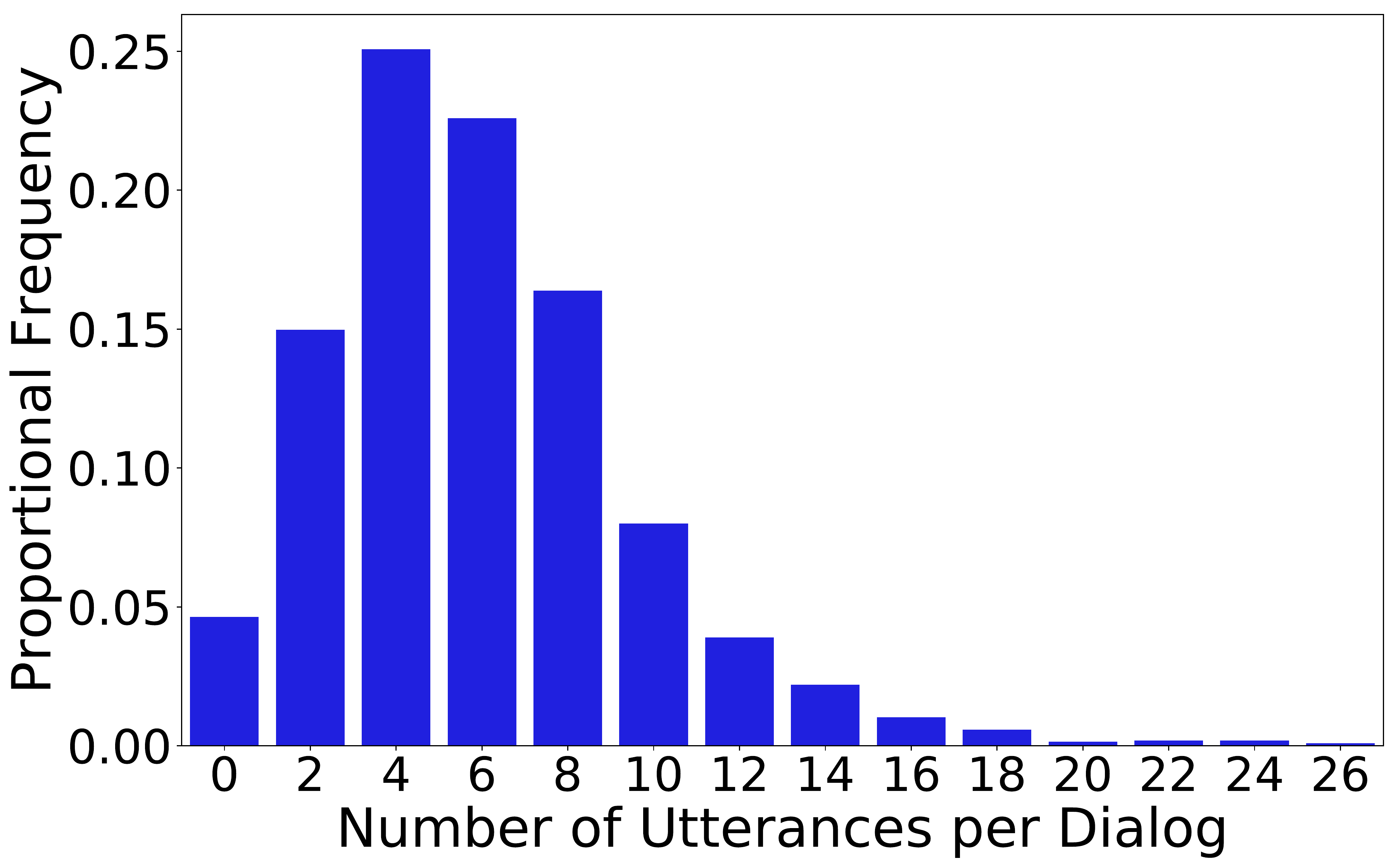}
\end{tabular}
\caption{The distributions of steps taken by human \nav{}s versus a shortest path planner (Left), the number of word tokens from the \nav{} and the \ora{} (Center), and the number of utterances in dialogs across the \dataset{} dataset.}
\label{fig:steps_and_words}
\vspace{-4mm}
\end{figure}

\paragraph{Dialog Prompts.}
A dialog prompt is a tuple of the house scan $S$, a target object $t_o$ to be found, a starting position $p_0$, and a goal region $G_j$.
We use the MatterPort object segmentations to get region locations for household objects, as in prior work~\cite{nguyen:cvpr19}.
We define a set of 81 unique object types that appear in at least 5 unique houses and appear between 2 and 4 times per such house.\footnote{We also cut odd (``soffet'') and non-specific (``wall'') objects, and merge similar object names (e.g., ``potted plant'' and ``plant'') to cut down the initial 929 object types to these salient 81. Some houses do not have objects that meet our criteria, so \dataset{} represents only 83 of the 90 total MatterPort houses.}
Each dialog begins with a hint, such as ``The goal room contains a \textit{plant},'' which by construction is both ambiguous (there are two to four rooms with a plant) and underspecified (the path to the room is not described by the hint).

Given a house scan $S$ and a target object $t_o$, a dialog prompt is created for every goal region $G_j$ in the house containing an instance of $t_o$.
Goal regions are sets of nodes that occupy the same room in a house scan.
The starting node $p_0$ is chosen to maximize the distance between $p_0$ and the goal regions $G_{0:|G|}$ containing $t_o$.
Formally,
\begin{equation*}
    p_0 = \text{argmax}_{p\in S}\left(\sqrt{\sum_{j}{\min_{p_i\in G_j}(d_h(p, p_i)^2)}}\right).
\end{equation*}

\paragraph{Crowdsourced Data Collection.}
We gathered human-human dialogs through Amazon Mechanical Turk.\footnote{\url{https://cvdn.dev/}. Connect with two tabs to start a dialog with yourself.}
In each Human Intelligence Task (HIT), workers read about the roles of \nav{} and \ora{} and could practice using the navigation interface.
Pairs of workers were connected to one another via a chat interface.

Every dialog was instantiated via a randomly chosen prompt $(S, t_o, p_0, G_j)$, with the \nav{} starting at panorama $p_0$ and both workers instructed via the text: ``Hint: The goal room contains a $t_o$.''
The dialog begins with the \nav{}'s turn.
On the \nav{}'s turn, they could navigate, type a natural language question to ask the \ora{}, or guess that they had found the goal room.
Incorrect guesses disabled further navigation and forced the \nav{} to ask a question to the \ora{}.
Throughout navigation, the \ora{} was shown the steps being taken as a mirror of the \nav{}'s interface, so that both workers were always aware of the current visual frame.
On the \ora{}'s turn, they could view an animation depicting the next $5$ hops through the navigation graph towards the goal room according to a shortest path planner and communicate back to the \nav{} via natural language (Figure~\ref{fig:full_demo}).
Five hops was chosen because this is slightly shorter than the $6$ hop average path in the R2R dataset, for which human annotators were able to provide reasonable language descriptions.
Each HIT paid $\$1.25$ per worker, the entire dataset collection cost over \$7k.

After successfully locating the goal room, workers rated their partner's cooperativeness (from 1 to 5).
Workers who failed to maintain a 4 or higher average peer rating were disallowed from taking more of our HITs.
On average, dialog participants' mean peer rating is $4.52$ out of 5 across \dataset{}.

\paragraph{Analysis.}
The \dataset{} dataset has longer routes and language contexts than the R2R task.
The dialogs exhibit complex phenomena that require both dialog and navigation history to resolve.

\begin{table}[ht]
\centering
\begin{small}
\begin{tabular}{p{1.4cm}rrrp{8.5cm}}
    & \textbf{\textit{Dia}} & \textbf{\textit{Nav}} & \textbf{\textit{Ora}} & \textbf{Example} \\
    \toprule
    Ego & $92.5$ & $52.9$ & $65.8$ & \ora{}: Turn slightly to {\color{blue}your right} and go {\color{blue}forward} down the hallway \\
    \cmidrule{5-5}
    Needs Q & $13.0$ & - & $3.9$ & \nav{}: Should I turn left down the hallway ahead? \\
    & & & &  \ora{}: {\color{blue}ya} \\
    \cmidrule{5-5}
    Needs Dialog History & $3.5$ & $0.4$ & $1.0$ & \ora{}: Through the lobby. So go through the door next to the green towel. Go to the left door next to {\color{blue}the two yellow lights}. Walk straight to the end of the hallway and stop \\
    & & & & $\dots$ \\
    & & & & \nav{}: Are these {\color{blue}the yellow lights} you were talking about? \\
    \cmidrule{5-5}
    Needs Nav History & $14.0$ & $1.5$ & $3.4$ & \ora{}: {\color{blue}You were there briefly but left}. There is a turntable behind you a bit. Enter the bedroom next to it. \\
    \cmidrule{5-5}
    Repair & $12.5$ & $1.6$ & $3.4$ & \ora{}: I am so sorry {\color{blue}I meant for you to look over to the right not the left} \\
    \cmidrule{5-5}
    Off-topic & $3.0$ & $5.4$ & $5.1$ & \nav{}: I am to the `rear' of the zebra. {\color{blue}Nice one.} \\
    & & & & \ora{}: {\color{blue}Ok hold your nose} and go to the left of the zebra, through the livingroom and kitchen and towards the bedroom you can see past that \\
    \cmidrule{5-5}
    Vacuous & $6.0$ & $22.7$ & $2.3$ & \nav{}: {\color{blue}Ok, now where?} \\
    \bottomrule \\
\end{tabular}
\end{small}
\caption{
The average percent of \textit{Dialogs}, as well as individual \nav{} and \ora{} utterances, exhibiting each phenomena out of 100 hand-annotated dialogs.
Two authors annotated each dialog and reached an agreement of Cohen's~$\kappa=.738$ across all phenomena labels.
}
\vspace{-8mm}
\label{tab:analysis}
\end{table}

Figure~\ref{fig:steps_and_words} shows the distributions of path lengths, word counts, and number of utterances across dialogs in the \dataset{} dataset.
Human ($25.0\pm12.9$) and planner ($17.4\pm7.0$) path lengths are on average more than three times longer, and have higher variance, than the path lengths in R2R ($6.0\pm0.85$).
Average word counts for navigators ($33.5$) and oracles ($48.1$) sum to an average $81.6$ words per dialog, again exceeding the Room-to-Room average of $29$ words per instruction by nearly a factor of three.
Dialogs average about 6 utterances each (3 question and answer exchanges), with a fraction being much longer---up to 26 utterances.
Some dialogs have no exchanges (about 5\%): the \nav{} was able to find the goal location by intuition alone given the hint.
Because more than one room always contains $t_o$, these are `lucky' guesses.

We randomly sampled 100 dialogs with at least one QA exchange and annotated whether each utterance (out of 342 per speaker) exhibited certain phenomena (Table~\ref{tab:analysis}).
Over half the utterances from both \nav{} and \ora{} roles, and over 90\% of all dialogs, contain egocentric references requiring the agent's position and orientation to interpret.
Some \ora{} answers require the \nav{} question to resolve (e.g., when the answer is just a confirmation).
Some utterances need dialog history from previous exchanges or past visual navigation information.
More than 10\% of dialogs exhibit conversational repair, when speakers try to rectify mistakes.
Speakers sometimes establish rapport with off-topic comments and jokes.
Both speakers, especially those in the \nav{} role, sometimes send vacuous communications, but this is limited to a smaller percentage of dialogs.

Models attempting to perform navigation, ask questions, or answer questions about an embodied environment must grapple with these types of phenomena.
For example, an agent may need to attend not just to the last QA exchange, but to the entire dialog and navigation history in order to correctly follow instructions.

%% file: 04_task.tex
\dataset{} facilities training agents for navigation, question asking, and question answering.
In this paper, we focus on navigation.
The ability to navigate successfully given dialog history is key to any future work in the Vision-and-Dialog Navigation paradigm.
Every dialog is a sequence of \nav{} question and \ora{} answer exchanges, with \nav{} steps following each exchange.
We use this structure to divide dialogs into \taskfull{} (\task{}) instances.

In particular, \dataset{} instances are each comprised of a repeating sequence ${<N_0, Q_1, A_1, N_1, \dots, Q_k, A_k, N_k>}$ of navigation actions, $N$, questions asked by the \nav{}, $Q$, and answers from the \ora{}, $A$.
Because sending a question or answer ends the worker's turn, every question $Q_i$ and answer $A_i$ is a single string of tokens.
For each dialog with prompt $(S, t_o, p_0, G_j)$, an \task{} instance is created for each of $0\leq i\leq k$.
The input is $t_o$ and a (possibly empty) history of questions and answers $(Q_{1:i}, A_{1:i})$.
The task is to predict navigation actions that bring the agent closer to the goal location $G_j$, starting from the terminal node of $N_{i-1}$ (or $p_0$, for $N_0$).
We extract 7415 \task{} instances from the 2050 navigation dialogs in \dataset{}.

\begin{figure}[ht]
\centering
\includegraphics[width=1.\columnwidth]{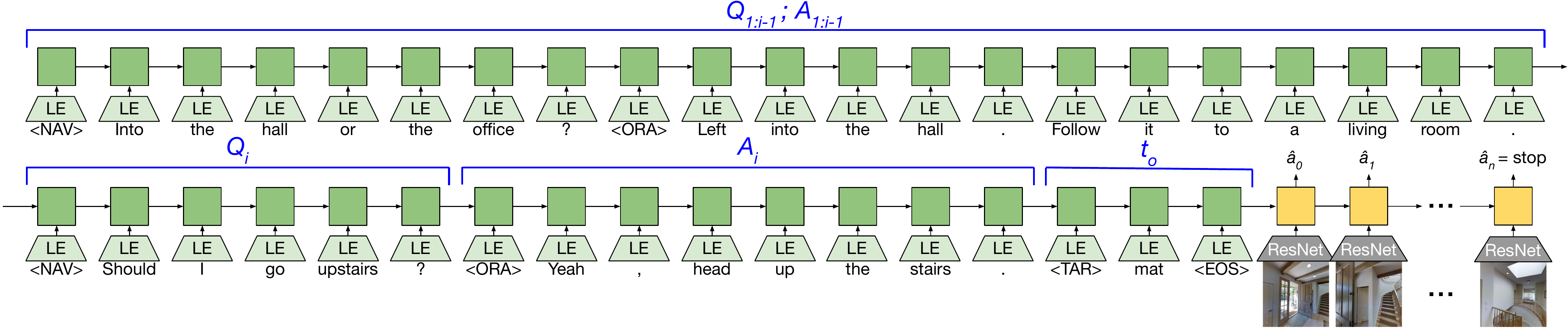}
\caption{We use a sequence-to-sequence model with an LSTM encoder that takes in learnable token embeddings (LE) of the dialog history.
The encoder conditions an LSTM decoder for predicting navigation actions that takes in fixed ResNet embeddings of visual environment frames.
Here, we demarcate subsequences in the input (e.g., $t_o$) compared during input ablations.
}
\label{fig:model}
\vspace{-3mm}
\end{figure}

We divide these instances into training, validation, and test folds, preserving the R2R folds by house scan.
This division is further done by dialog, such that for every dialog in \dataset{} the \task{} instances created from it all belong to the same fold.
As in R2R, we split the validation fold into \textit{seen} and \textit{unseen} house scans, depending on whether the scan is present in the training set. This results in 4742 training, 382 \textit{seen} validation, 907 \textit{unseen} validation, and 1384 \textit{unseen} test instances.

We provide two forms of supervision for the \task{} task: $N_i$, the navigation steps taken by the \nav{} after question-answer exchange $i$, and $O_i$, the shortest-path steps shown to the \ora{} and used as context to provide answer $A_i$.
In each instance of the task, $i$ indexes the QA exchange in the dialog from which the instance is drawn (with $i=0$ an empty QA followed by initial navigation steps).
Across \task{} instances, the $N_i$ steps range in length from 1 to 40 (average $6.63$), and the $O_i$ steps range in length from 0 to 5 (average $4.35$).
The \nav{} often continues farther than what the \ora{} describes, using their intuition about the house layout to seek the target object.

We evaluate performance on this task by measuring how much progress the agent makes towards $G_j$.
Let $e(P)$ be the end node of path $P$, $b(P)$ the beginning, and $\hat{P}$ the path inferred by the navigation agent.
Then the progress towards the goal is defined as the reduction (in meters) from the distance to the goal region $G_j$ at $b(\hat{P})$ versus at $e(\hat{P})$.
Because $G_j$ is a set of nodes, we take the minimum distance $\min_{p\in G_j}(d_m(p, q))$ as the distance between $q$ and region $G_j$.
Note that this is a topological distance (e.g., we measure the distance around a wall, rather than straight through it).

%% file: 05_experiments.tex
\citet{anderson:cvpr18} introduced a sequence-to-sequence model to serve as a learning baseline in the R2R task.
We formulate a similar model to encode an entire dialog history, rather than a single navigation instruction, as an initial learning baseline for the \task{} task.
The dialog history is encoded using an LSTM and used to initialize the hidden state of an LSTM decoder whose observations are visual frames from the environment, and whose outputs are actions in the environment (Figure~\ref{fig:model}).

We replace words that occur fewer than 5 times with an \texttt{UNK} token.
The resulting vocabulary sizes are 1042 language tokens in the training fold and 1181 tokens in the combined training and validation folds.
We also use special \texttt{NAV} and \texttt{ORA} tokens to preface a speaker's tokens, \texttt{TAR} to preface the target object token, and \texttt{EOS} to indicate the end of the input sequence.
During training, an embedding is learned for every token and given as input to the encoder LSTM.
For visual features, we embed the visual frame as the penultimate layer of an Imagenet-pretrained ResNet-152 model~\cite{he:cvpr16}.

When evaluating against the validation folds, we train only on the training fold.
When evaluating against the test fold, we train on the union of the training and validation folds.
We ablate the distance of dialog history encoded, and introduce a mixed planner and human supervision strategy at training time.
We hypothesize both that encoding a longer dialog history and using mixed-supervision steps will increase the amount the agent progresses towards the goal.

\begin{table}[ht]
\centering
\begin{small}
\begin{tabular}{ccccccc>{\raggedleft\arraybackslash}p{1.5cm}>{\raggedleft\arraybackslash}p{1.5cm}>{\raggedleft\arraybackslash}p{1.5cm}}
    & & \multicolumn{5}{c}{\textbf{Seq-2-Seq Inputs}} & \multicolumn{3}{c}{\textbf{Goal Progress} (m) $\uparrow$} \\
    & & & & & & $Q_{1:i-1} $ & & & \\
    \textbf{Fold} & & $V$ & $t_o$ & $A_i$ & $Q_i$ & $A_{1:i-1}$ & \textbf{Oracle} & \textbf{Navigator} & \textbf{Mixed} \\
\toprule
    \multirow{9}{*}{\rotatebox[origin=c]{90}{Val (Seen)}} & \multirow{5}{*}{\rotatebox[origin=c]{90}{Baselines}} & \multicolumn{5}{c}{\texttt{Shortest Path} Agent} & $8.29$ & $7.63$ & $9.52$ \\
    & & \multicolumn{5}{c}{\texttt{Random} Agent} & $0.42$ & $0.42$ & $0.42$ \\
    & & & & & & & $0.59$ & $0.83$ & $0.91$ \\
    & & \cblkmark & & & & & $4.12$ & $5.58$ & $5.72$ \\
    & & & \cblkmark & \cblkmark  & \cblkmark  & \cblkmark & $1.41$ & $1.43$ & $1.58$ \\
    \cmidrule{2-10}
    & \multirow{4}{*}{\rotatebox[origin=c]{90}{Ours}} & \cblkmark & \cblkmark & & & & $4.16$ & \good{$\pmb{5.71}$} & \good{$5.71$} \\
    & & \cblkmark & \cblkmark & \cblkmark & & & $4.34$ & $5.61$ & \good{$6.04$} \\
    & & \cblkmark & \cblkmark & \cblkmark & \cblkmark & & $4.28$ & $5.58$ & \good{$\pmb{6.16}$} \\
    & & \cblkmark & \cblkmark & \cblkmark & \cblkmark & \cblkmark & $\pmb{4.48}$ & $5.67$ & \good{$5.92$} \\
    \midrule
    \multirow{9}{*}{\rotatebox[origin=c]{90}{Val (Unseen)}} & \multirow{5}{*}{\rotatebox[origin=c]{90}{Baselines}} & \multicolumn{5}{c}{\texttt{Shortest Path} Agent} & $8.36$ & $7.99$ & $9.58$ \\
    & & \multicolumn{5}{c}{\texttt{Random} Agent} & $1.09$ & $1.09$ & $1.09$ \\
    & & & & & & & $0.69$ & $1.32$ & $1.07$ \\
    & & \cblkmark & & & & & $0.85$ & $1.38$ & $1.15$ \\
    & & & \cblkmark & \cblkmark & \cblkmark & \cblkmark & $1.68$ & $1.39$ & $1.64$ \\
    \cmidrule{2-10}
    & \multirow{4}{*}{\rotatebox[origin=c]{90}{Ours}} & \cblkmark & \cblkmark & & & & $0.74$ & \good{$1.33$} & $1.29$ \\
    & & \cblkmark & \cblkmark & \cblkmark & & & $1.14$ & $1.62$ & \good{$2.05$} \\
    & & \cblkmark & \cblkmark & \cblkmark & \cblkmark & & $1.11$ & $1.70$ & \good{$1.83$} \\
    & & \cblkmark & \cblkmark & \cblkmark & \cblkmark & \cblkmark & $\pmb{1.23}$ & $\pmb{1.98}$ & \good{$\pmb{2.10}$} \\
    \midrule
    \multirow{9}{*}{\rotatebox[origin=c]{90}{Test (Unseen)}} & \multirow{5}{*}{\rotatebox[origin=c]{90}{Baselines}} & \multicolumn{5}{c}{\texttt{Shortest Path} Agent} & $8.06$ & $8.48$ & $9.76$ \\
    & & \multicolumn{5}{c}{\texttt{Random} Agent} & $0.83$ & $0.83$ & $0.83$ \\
    & & & & & & & $0.13$ & $0.80$ & $0.52$ \\
    & & \cblkmark & & & & & $0.99$ & $1.56$ & $1.74$ \\
    & & & \cblkmark & \cblkmark  & \cblkmark  & \cblkmark  & $1.51$ & $1.20$ & $1.40$ \\
    \cmidrule{2-10}
    & \multirow{4}{*}{\rotatebox[origin=c]{90}{Ours}} & \cblkmark & \cblkmark & & & & $1.05$ & $1.81$ & \good{$1.90$} \\
    & & \cblkmark & \cblkmark & \cblkmark & & & $1.21$ & $2.01$ & \good{$2.05$} \\
    & & \cblkmark & \cblkmark & \cblkmark & \cblkmark & & $\pmb{1.35}$ & $1.78$ & \good{$2.27$} \\
    & & \cblkmark & \cblkmark & \cblkmark & \cblkmark & \cblkmark & $1.25$ & $\pmb{2.11}$ & \good{$\pmb{2.35}$} \\
    \bottomrule \\
\end{tabular}
\end{small}
\caption{
Average agent progress towards the goal location when trained using different path end nodes for supervision.
Among sequence-to-sequence ablations, \textbf{bold} indicates most progress across available language input, and \good{blue} indicates most progress across supervision signals.
}
\vspace{-7mm}
\label{tab:navigation}
\end{table}

\paragraph{Training.}
Given supervision from an end node $e(P^*)$, the agent infers navigation actions to form path $\hat{P}$.
We train all agents with student-forcing for 20000 iterations of batch size 100, and evaluate validation performance every 100 iterations (see the Appendix for details).
The best performance across all epochs is reported for validation folds.
At each timestep the agent executes its inferred action $\hat{a}$, and is trained using cross entropy loss against the action $a^*$ that is next along the shortest path to the end node $e(P^*)$.
Using the whole navigation path, $P^*$, as supervision rather than only the end node has been considered in other work~\cite{jain:acl19}.
At test time, the agents are trained up to the epoch that achieved the best performance on the \textit{unseen} validation fold and then evaluated (e.g., test fold evaluations are run only \textit{once} per agent).

Recall that for each \task{} instance, the path shown to the \ora{} during QA exchange $i$, $O_i$, and the path taken by the \nav{} after that exchange, $N_i$, are given.
We define the mixed supervision path $M_i$ as $N_i$ when $e(O_i)\in N_i$, and $O_i$ otherwise.
This new form of supervision has parallels to previous works on learning from imperfect or adversarial human demonstrations.
One common solution is to use imperfect human demonstrations to learn an initial policy which is then refined with Reinforcement Learning (RL)~\cite{taylor2011integrating}.
Learning performance can be improved by first assigning a confidence measure to the demonstrations and only including those demonstrations that pass a certain threshold~\cite{wang2017improving}.
While we leave the evaluation of more sophisticated RL methods to future work, the mixed supervision described above can be thought of as using a simple binary confidence heuristic to threshold the human demonstrations.

\paragraph{Baselines and Ablations.}
We compare the sequence-to-sequence agent to a full-state information shortest path agent, to a non-learning baseline, and to unimodal baselines.
The \texttt{Shortest Path} agent takes the shortest path to the supervision goal at inference time, and represents the best a learning agent could do under a given form of supervision.
The non-learning \texttt{Random} agent chooses a random heading and walks up to 5 steps forward (as in~\cite{anderson:cvpr18}).
Random baselines can be outperformed by unimodal model ablations---agents that consider only visual input, only language input, or neither---on VLN tasks~\cite{thomason:naacl19}.
So, we also compare our agent to unimodal baselines where agents have zeroed out visual features in place of the $V$ ResNet features at each decoder timestep (vision-less baseline) and/or empty language inputs to the encoder (language-less baseline).
To examine the impact of dialog history, we consider agents with access to the target object $t_o$; the last \ora{} answer $A_i$; the prefacing \nav{} question $Q_i$; and the full dialog history (Figure~\ref{fig:model}).

\paragraph{Results.}
Table~\ref{tab:navigation} shows agent performances given different forms of supervision.
We ran paired $t$-tests between all model ablations within each supervision paradigm and across paradigms, and applied the Benjamini--Yekutieli procedure to control the false discovery rate (details in the Appendix).

Using all dialog history significantly outperforms unimodal ablations in \textit{unseen} environments.
The \texttt{Shortest Path} agent performance with \nav{} supervision $N_i$ approximates human performance on \task{}, because $e(N_i)$ is the node reached by the human \nav{} after QA exchange $i$ during data collection.
The sequence-to-sequence models establish an initial, multimodal baseline for \task{}, with headroom remaining compared to human performance, especially in \textit{unseen} environments.
Using all dialog history, rather than just the last question or question-answer exchange, is needed to achieve statistically significantly better performance than using the target object alone in \textit{unseen} test environments.
This supports our hypothesis that dialog history is beneficial for understanding the context of the latest navigation instruction $A_i$.
Models trained with mixed supervision always statistically significantly outperform those trained with oracle or navigator supervision.
This supports our hypothesis that using human demonstrations only when they appear trustworthy increases agent progress towards the goal.

%% file: 06_conclusion.tex
We introduce \datasetfull{}: 2050 human-human, situated navigation dialogs in a photorealistic, simulated environment.
The dialogs contain complex phenomena that require egocentric visual grounding and referring to both dialog history and past navigation history for context.
\dataset{} is a valuable resource for studying \textit{in-situ} navigation interactions, and for training agents that both navigate human environments and ask questions when unsure, as well as those that provide verbal assistance to humans navigating in unfamiliar places.

We then define the \taskfull{} task.
Our evaluations show that dialog history is relevant for navigation agents to learn a mapping between dialog-based instructions and correct navigation actions.
Further, we find that using a mixed form of both human and planner supervision combines the best of each: long-range exploration of an environment according to human intuition to find the goal, and short-range accuracy aligned with language input.

\paragraph{Limitations.}
The \dataset{} dataset builds on the Room-to-Room task in the MatterPort Simulator~\cite{anderson:cvpr18}.
We would like to use \dataset{} to train real world agents for dialog and navigation.
Simply fine-tuning on real world data may not be sufficient.
Real-world robot navigation relies on laser scan depths, not just RGB information, and invokes lower quality egocentric vision, sensor noise, and localization issues.
While the simulation provides photorealistic environments, it suffers from discrete, graph-based navigation, requiring a real world navigable environment to be mapped and divided into topological waypoints.
Human-human dialogs collected in high-fidelity, continuous motion simulators (e.g.,~\cite{ai2thor}) or using virtual reality technology may facilitate easier transfer to physical robot platforms.
However, sharing a simulation environment with the existing R2R task means that models for dialog history tasks like \task{} may benefit from pretraining on R2R.

\paragraph{Future Work.}
The sequence-to-sequence model used in our experiments serves as an initial learning baseline for the \task{} task.
Moving forward, by formulating \task{} as a sequential decision process we can use RL to shape the agent's policy, as in recent VLN work~\cite{tan:naacl19}.
Dialog analysis also suggests that there is relevant information in the historical navigation actions which are not considered by the initial model.
Jointly conditioning dialog and navigation history may help resolve past reference instructions like ``Go back to the stairwell and go up one flight of steps,'' and could involve cross-modal attention alignment.

The \dataset{} dataset also provides a scaffold for navigation-centered question asking and question answering tasks.
In our future work, we will explore training two agents in tandem: one to navigate and ask questions when lost, and another to answer those questions.
This will facilitate end-to-end evaluation on \dataset{}, and will differ from all existing VLN tasks by involving two, trained agents engaged in task-oriented dialog.

%% file: 07_appendix.tex
\subsection{Additional \dataset{} Analysis}

\begin{figure}[ht]
\centering
\includegraphics[width=0.9\columnwidth]{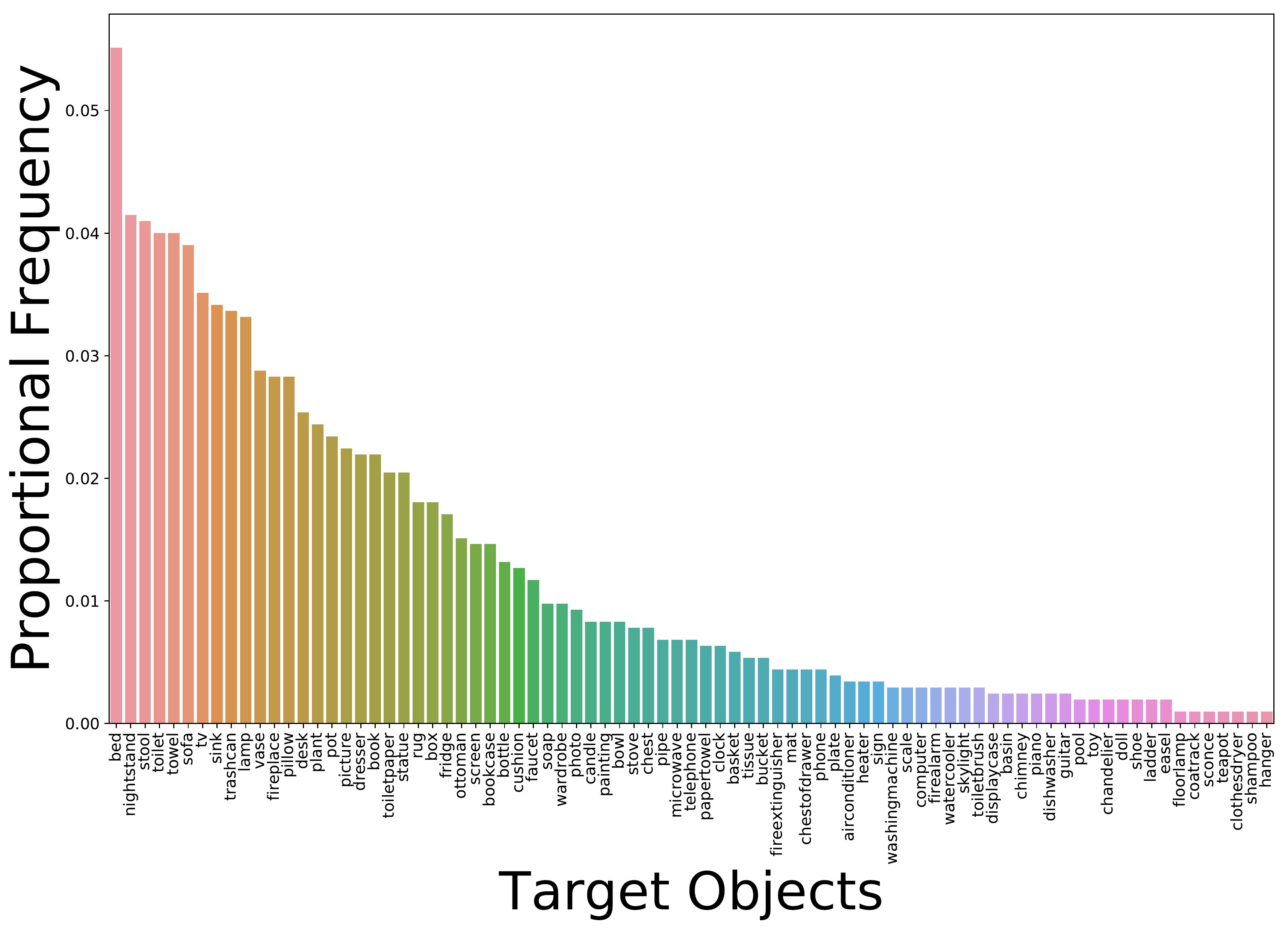}
\caption{The distribution of the 81 target objects $t_o$ in dialogs across \dataset{}.}
\label{fig:object_dist}
\end{figure}

Figure~\ref{fig:object_dist} gives the distributions of target objects $t_o$ across the dialogs in \dataset{}.
The most frequent objects are those that are both frequent across houses and typically number between 2 and 4 per house, and often have a one-to-one correspondence with bedrooms and bathrooms.

\begin{figure}[ht]
\begin{tabular}{cc}
    \includegraphics[width=0.45\columnwidth]{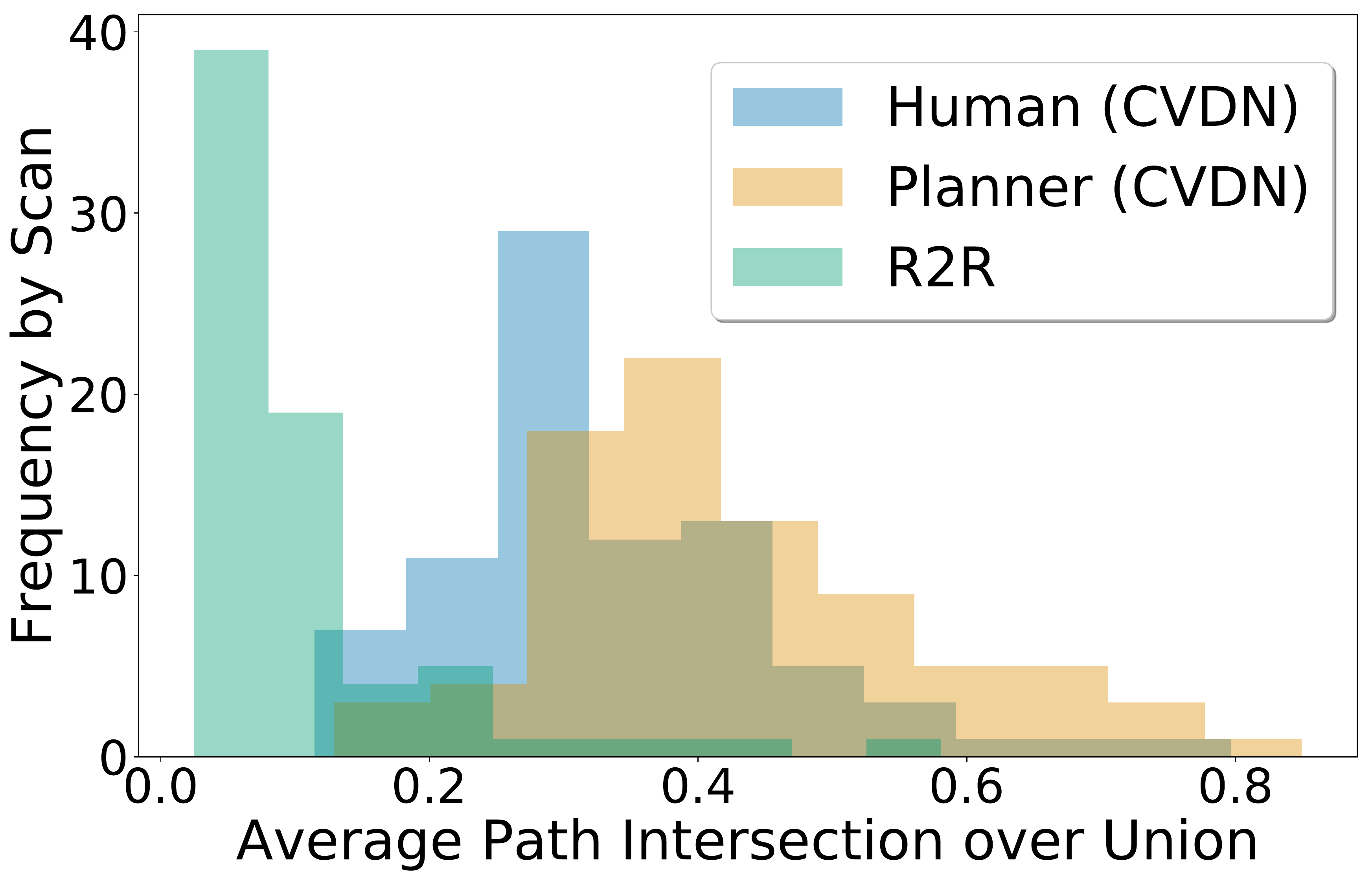} &
    \includegraphics[width=0.45\columnwidth]{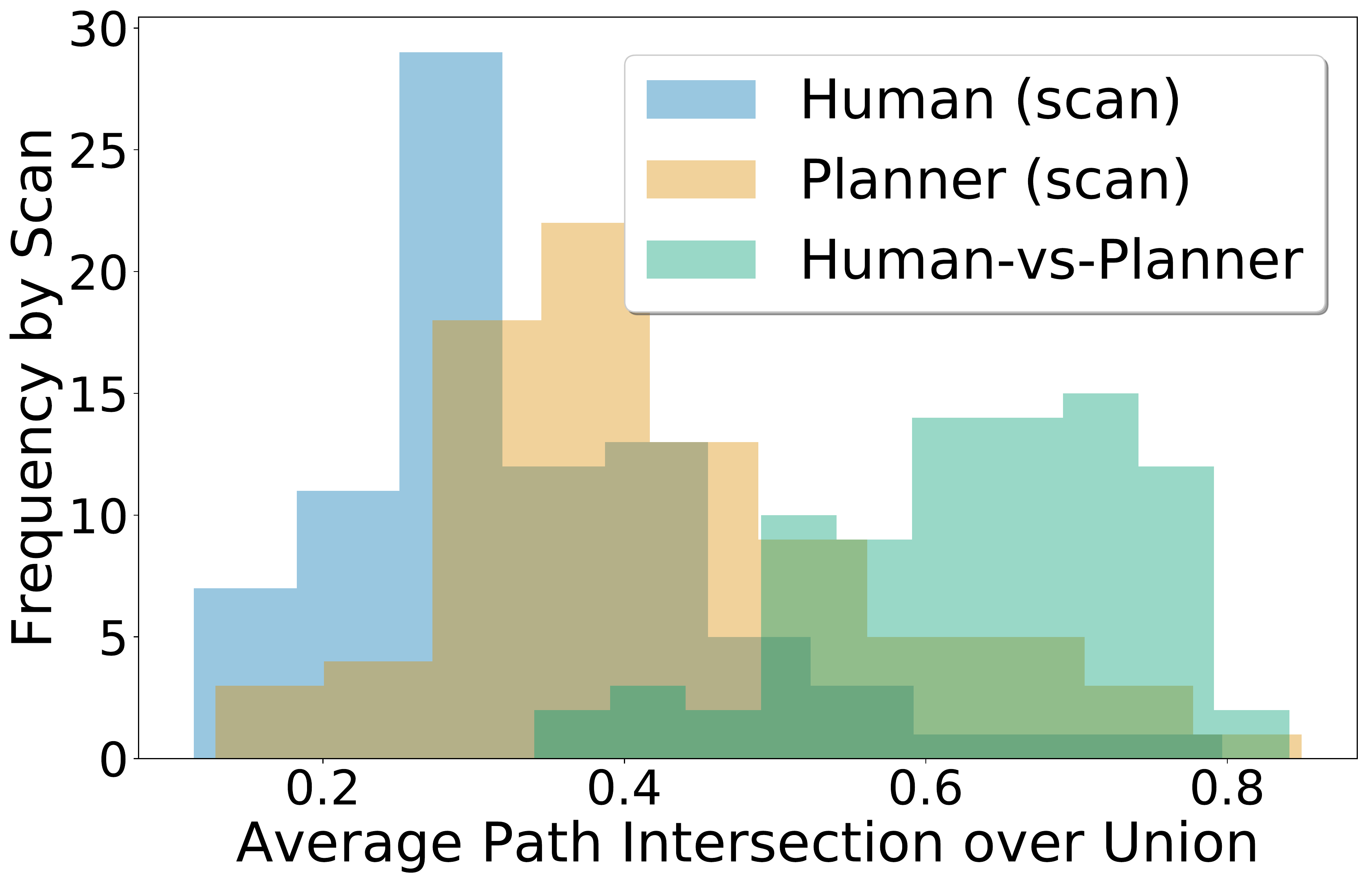}
\end{tabular}
\caption{Left: The IoU of nodes in the paths of human \nav{} and shortest path planner trajectories in \dataset{} versus those in R2R when comparing paths in the same scan.
Right: The IoU of \nav{} and shortest path planner trajectories in the same scan versus the IoU of player and shortest path planner trajectories across a dialog.}
\label{fig:iou_comp}
\end{figure}

Figure~\ref{fig:iou_comp} gives the intersection-over-union (IoU) of paths in \dataset{} within the same scan, comparing them against those in R2R and human performance per-dialog.
The average path IoU across a scan is the average number of navigation nodes in the intersection of two paths over the union of nodes in those paths, across all paths in the scan.
Compared to R2R, the paths in the dialogs of \dataset{} share more navigation nodes per scan because of the way starting panoramas $p_0$ were chosen---to maximize the distance to potential goal regions.
Many \dataset{} paths start at or near the same remote $p_0$ nodes in, e.g., basements, rooftops, and lawns.
Per-dialog, we measure the IoU between human \nav{} and shortest path planner trajectories and find that there is substantially more overlap than between two paths in the same scan, indicating that humans follow closer to the shortest path than to an average walk through the scan (e.g., they are not just memorizing previous dialog trajectories).

\subsection{Additional \task{} Analysis}

\begin{figure}[ht]
\begin{tabular}{cc}
    \includegraphics[width=0.45\columnwidth]{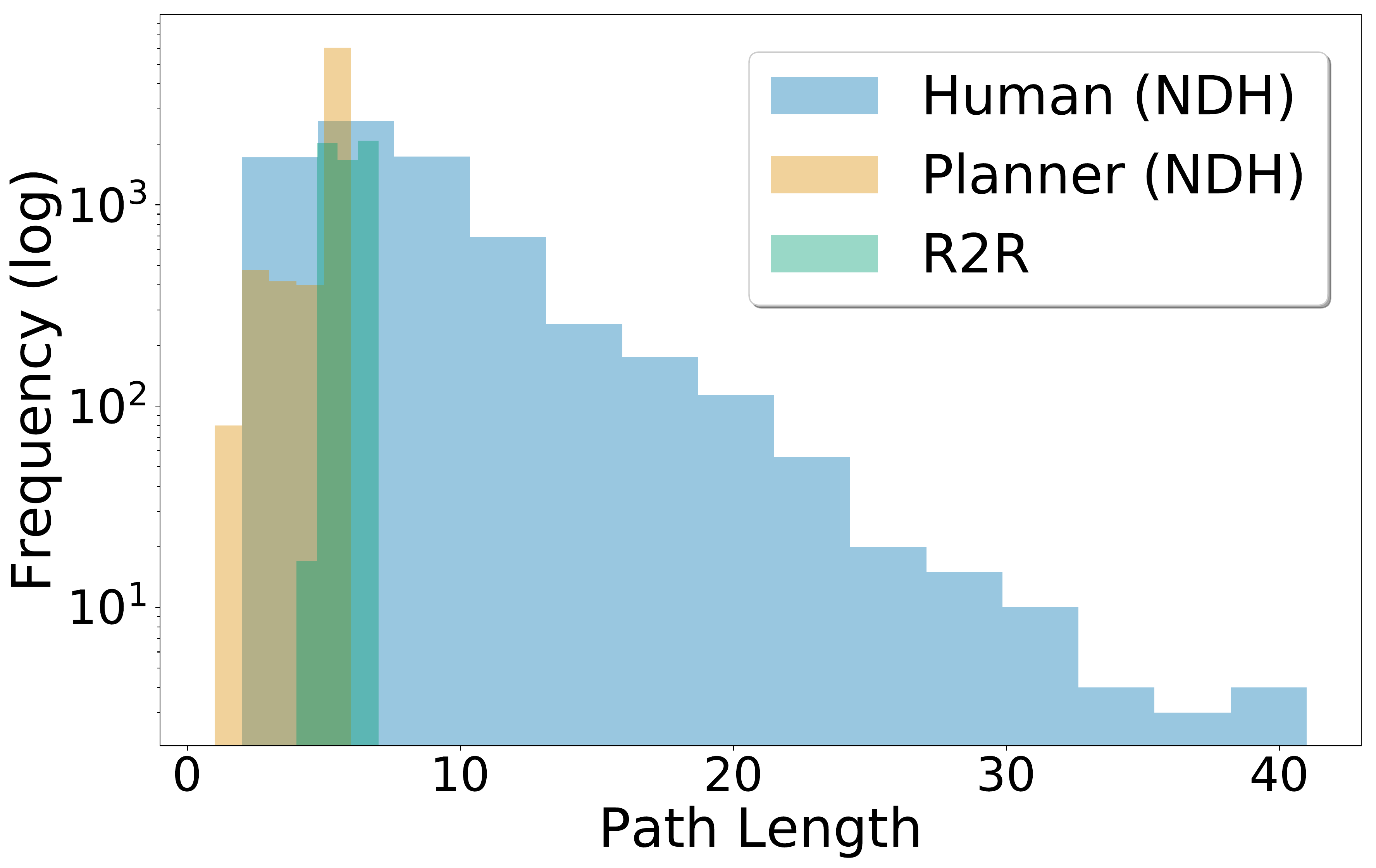} &
    \includegraphics[width=0.45\columnwidth]{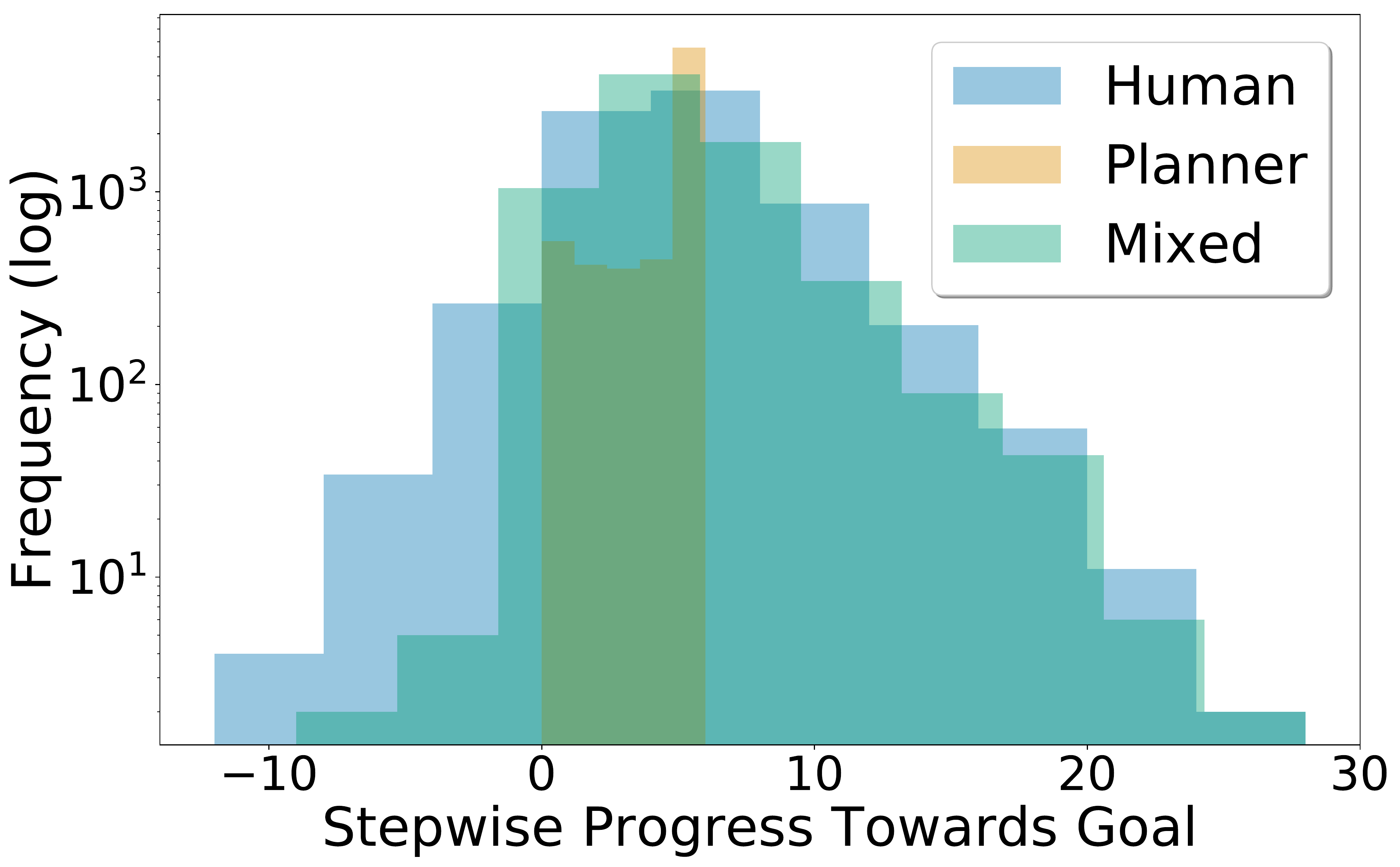}
\end{tabular}
\caption{Left: The distributions of path lengths by human \nav{} and the shortest path planner provided as supervision in \task{} instances versus path lengths in R2R supervision.
Right: The progress per NDH instance made towards the goal (in steps) by the human \nav{}, the shortest path planner, and the mixed-supervision path.}
\label{fig:ndh_dists}
\end{figure}

Figure~\ref{fig:ndh_dists} gives path data for the \task{} task.
Compared to R2R, path lengths using shortest path supervision ($O_i$) are on average shorter than those in R2R, because paths shown to the \ora{} were at most length 5.
By contrast, human \nav{} paths ($N_i$) are substantially longer than those seen in R2R.
We also examine the distribution of the number of hops progressed towards the goal per \task{} instance across \ora{} shortest path, human navigator, and mixed supervision ($M_i$).
While the planner always moves towards the goal (or stands still, if the \nav{} is already in the goal region), human \nav{}s sometimes move farther away from the goal, though in general make more progress than the planner.
Using mixed supervision, fewer trajectories move ``backwards''; the simple heuristic of whether a \nav{} walked over the last node in the \ora{}'s described shortest path shifts the distribution weight farther towards positive goal progress.

\subsection{\task{} Model Performance Statistical Comparisons}
We ran paired $t$-tests between all model ablations within each supervision paradigm (e.g., comparing all mixed supervision models to one another), and across paradigms (e.g., comparing the full dialog history model trained with mixed supervision to the one with navigator path supervision).
Data pairs are NDH the distances progressed towards the goal on the same instance (i.e., dialog history and goal) between two conditions.
This results in hundreds of comparison tests, so we apply a Benjamini--Yekutieli procedure to control the false discovery rate.
Because the tests are not all independent, but some are, we estimate $c(m)$ under an arbitrary dependence assumption as $c(m)~=~\sum_{i+1}^{m}{\frac{1}{i}}$, where $m$ is the number of tests run.
We choose a significance threshold of $\alpha<0.05$.
Rather than report the hundreds of individual $p$-values, we highlight salient results below.

\paragraph{Different forms of supervision}
With one exception, in all environments (\emph{seen} validation, \emph{unseen} validation, and \emph{unseen} test), across ablations of language context (i.e., full model using all history down to model using only the target object as dialog context), the differences in progress towards the goal under oracle, navigator, and mixed supervision are statistically significantly different.
The only exception is the difference between oracle and navigator path supervision in \emph{unseen} validation environments with the last answer only (i.e., row 16 of Table~\ref{tab:navigation}) ($p=0.006$).
Models trained with mixed supervision almost always achieve the most progress. For brevity, below we discuss further comparisons between models trained with mixed supervision.

\paragraph{Different amounts of dialog history}
In \emph{unseen} validation and test environments, using all dialog history statistically significantly outperforms using only the target object, but not only the last answer ($p=0.773$ in validation, $p=0.035$ in test) or the last question-answer exchange ($p=0.143$ in validation, $p=0.560$ in test).
Notably, in test environments, a statistically significant difference compared to using only the target object is observed \emph{only} when using all dialog history.
In \emph{seen} validation houses, adding additional dialog history does not result in statistically significant gains, reflecting the representative power of the vision-only unimodal baseline.

\paragraph{Unimodal ablations}
In \emph{unseen} validation and test environments, the model using all dialog history statistically significantly outperforms the unimodal baselines in all cases except the language-only unimodal model in the \emph{unseen} validation houses ($p=0.011$).
In \emph{seen} validation houses, this model statistically significantly outperforms the language-only and zero (no language, no vision) unimodal ablations.
This result does not hold for the vision-only baseline, which is able to memorize the familiar houses for use at test time.

\subsection{Sequence-to-Sequence Model Training}

\paragraph{Hyperparameters.}
We use the training hyperparameters (optimizer, learning rate, hidden state sizes, etc.) presented in \citet{anderson:cvpr18} when training our sequence-to-sequence agents.
We adjust the maximum input sequence length for language encoding based on the amount of dialog history available: 3 for $t_o$ only (e.g., \texttt{TAR} tag, the target itself, and \texttt{EOS}); 70 for $A_i$; 120 for adding $Q_i$; and 720 (e.g., 120 times 6 turns of history) for $Q_{1:k},A_{1:k}$.
We increase the maximum episode length (e.g., the maximum number of navigation actions) depending on the supervision being used: 20 for oracle $O_i$ (the same as in R2R) and 60 for navigator $N_i$ and mixed $M_i$.

\paragraph{Teacher- versus Student-Forcing.}
We use student-forcing when training all of our sequence-to-sequence agents.
\citet{anderson:cvpr18} found that student-forcing improved agent performance in unseen environments.
Further, \citet{thomason:naacl19} found that agents trained via teacher-forcing were outperformed by their unimodal ablations (i.e., they did not learn to incorporate both language and vision supervision, instead memorizing unimodal priors).
Thus, we see no value in evaluating multi-modal agents trained via teacher-forcing in this setting.

\paragraph{Language Encoding.}
It is common in sequence-to-sequence architectures to reverse the input sequence of tokens during training, because the tokens relevant for the first decoding actions are likely also the first in the input sequence.
Reversing the sequence means those relevant tokens have been seen more recently by the encoder, and this strategy was employed in prior work~\cite{anderson:cvpr18}.
Following this intuition, we preserve the order of the dialog history during encoding, so that the most recent utterances are read just before decoding, but reverse the tokens at the utterance level (e.g., $Q_i$ in Figure~\ref{fig:model} is represented as sequence ``\texttt{<NAV>} ? upstairs go I Should'').

\subsection{Naive Dialog History Encoding}

We naively concatenated an encoded navigation history $N_H$ (via an LSTM taking in ResNet embeddings of past navigation frames) to the encoded dialog history, then learned a feed-forward shrinking layer to initialize the decoder (Table~\ref{tab:navigation_history_appendix}).
We hypothesize that there is some signal in this information, but we discover that naive concatenation does not improve performance in \textit{seen} or \textit{unseen} environments.
We suspect that a modeling approach which learns an attention alignment between the navigation history and dialog history could make better use of the additional signal.

\begin{table}[ht]
\centering
\begin{small}
\begin{tabular}{ccccccc>{\raggedleft\arraybackslash}p{1.5cm}>{\raggedleft\arraybackslash}p{1.5cm}>{\raggedleft\arraybackslash}p{1.5cm}}
    & \multicolumn{6}{c}{\textbf{Seq-2-Seq Inputs}} & \multicolumn{3}{c}{\textbf{Goal Progress} (m) $\uparrow$} \\
    & & & & & & $Q_{1:i-1} $ & & & \\
    \textbf{Fold} & $N_H$ & $V$ & $t_o$ & $A_i$ & $Q_i$ & $A_{1:i-1}$ & \textbf{Oracle} & \textbf{Navigator} & \textbf{Mixed} \\
    \toprule
    \multirow{3}{*}{\rotatebox[origin=c]{90}{Val (Se)}} & \multicolumn{6}{c}{\texttt{Shortest Path} Agent} & $8.29$ & $7.63$ & $9.52$ \\
    & & \cblkmark & \cblkmark & \cblkmark & \cblkmark & \cblkmark & $4.48$ & $5.67$ & $5.92$ \\
    & \cblkmark & \cblkmark & \cblkmark & \cblkmark & \cblkmark & \cblkmark & $4.47$ & $5.37$ & $5.82$ \\
    \midrule
    \multirow{3}{*}{\rotatebox[origin=c]{90}{Val (Un)}} & \multicolumn{6}{c}{\texttt{Shortest Path} Agent} & $8.36$ & $7.99$ & $9.58$ \\
    & & \cblkmark & \cblkmark & \cblkmark & \cblkmark & \cblkmark & $1.23$ & $1.98$ & $2.10$ \\
    & \cblkmark & \cblkmark & \cblkmark & \cblkmark & \cblkmark & \cblkmark & $1.19$ & $1.86$ & $1.84$ \\
    \bottomrule
\end{tabular}
\end{small}
\caption{
Average sequence-to-sequence agent performance when the agent encodes the entire navigation history $N_H$ compared against the \texttt{Shortest Path} upper bound and the agent encoding all dialog history across different supervision signals.
}
\label{tab:navigation_history_appendix}
\end{table}